\documentclass[conference, letterpaper]{IEEEtran}  
\IEEEoverridecommandlockouts 

\usepackage{cite}
\usepackage{amsmath, amssymb, amsfonts}
\usepackage{algorithmic}
\usepackage{algorithm}
\usepackage{graphicx}
\usepackage{textcomp}
\usepackage{xcolor}
\usepackage{booktabs}
\usepackage{multirow}
\usepackage{tabularx}
\usepackage{url}
\usepackage{balance} 
\usepackage{hyperref}
\hypersetup{
    colorlinks=true,
    linkcolor=black,
    citecolor=black,
    urlcolor=black,
    pdfborder={0 0 0},
}

\usepackage[font=small,labelfont=bf]{caption}

\title{Exploiting Vulnerabilities: Universal Adversarial Attacks on Vision-Language-Action Models in Robotics}

\author{
\normalsize Songhua Yang$^{1,2}$, Ziyu Liu$^{1}$, Yuanwei Liu$^{1}$, Xuetao Li$^{1,2}$, Xuanye Fei$^{3}$, He Huang$^{3}$, Zheng Wang$^{1}$, Miao Li$^{1,2,*}$
}

\begin{document}
\pdfminorversion=5
\pdfpagewidth=8.5in
\pdfpageheight=11in


\setlength{\topmargin}{0in}  
\setlength{\textheight}{9in}  
\setlength{\headheight}{0in}
\setlength{\headsep}{0in}

\nocite{*}
\maketitle
\enlargethispage{0.25in}  
\vspace*{0.25in}  

{
\renewcommand{\thefootnote}{}
\footnotetext{$^1$School of Computer Science, Wuhan University.}
\footnotetext{$^2$School of Robotics, Wuhan University.}
\footnotetext{$^3$School of Power and Mechanical Engineering, Wuhan University.}
\footnotetext{$\boldsymbol{*}$ Corresponding author: Miao Li (E-mail: limiao@whu.edu.cn).}
}
\begin{abstract}
Recently, Vision-Language-Action (VLA) models have revolutionized robotic manipulation by seamlessly integrating visual perception, language understanding, and action generation in an end-to-end learning framework. However, since these models are designed to interact directly with the physical world and humans, their security is critical, and even small vulnerabilities can lead to catastrophic failures.
In this work, we propose the Universal Adversarial Object, a sphere with optimized surface texture that significantly degrades task success rates when placed within the robot's field of view. Specifically, our approach introduces a multi-level attack framework that jointly disrupts trajectory planning, task execution, and action control.
We validate our method in both simulated and real-world robotic settings. Experimental results demonstrate that the adversarial object reduces the average task success rates by 31.2\%-39.9\% for two representative VLA models (Pi0 and RDT), with success rates dropping to near zero in complex scenarios.
\end{abstract}


\section{Introduction}
\label{sec:intro}
Vision-Language-Action (VLA) models have emerged as a transformative paradigm in robotics, enabling robots to seamlessly translate visual observations and natural language instructions into physical actions through end-to-end learning frameworks \cite{black2024pi, liurdt}. Unlike Large Language Models (LLMs) or Vision-Language Models (VLMs), which operate in the digital domain, VLA models directly control physical robots that manipulate objects and interact with humans \cite{ma2024survey, sapkota2025vision}. 
This unique capability significantly increases the security risks associated with VLA models. However, research on adversarial attacks and security evaluations for VLA is virtually non-existent, with almost all studies focused solely on performance improvements rather than addressing the critical security vulnerabilities these models face \cite{liu2025survey}.

\begin{figure}[!t]
    \centering
    \includegraphics[width=0.45\textwidth]{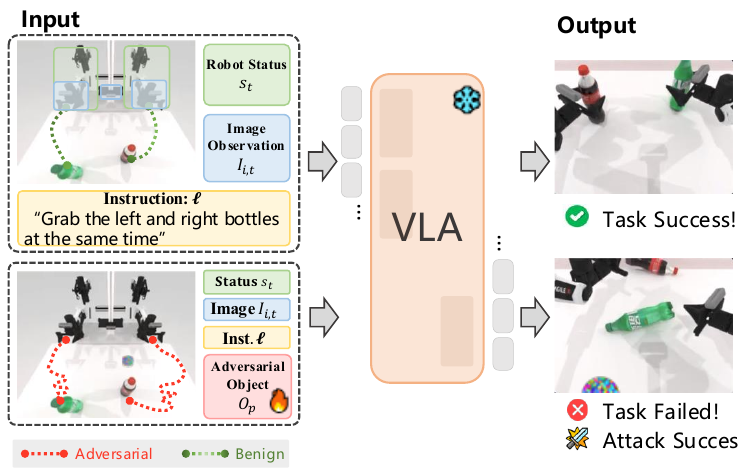}
    \caption{\textbf{An example of implementing an attack on VLA with universal adversarial objects.} The normal inputs to the VLA include the robot status $S_t$, image observation $I_{i,t}$, and instruction $l$, which enable the robot to successfully perform the grasping task. However, by applying an adversarial attack, an adversarial object $O_p$ is added to these normal inputs, causing the robot's grasping attempt to fail.}
    \label{fig1}
\end{figure}

Adversarial attack research serves as a critical tool for evaluating model robustness and identifying security vulnerabilities before deployment \cite{zhang2021survey}. Although physical adversarial attacks have successfully compromised traditional robotic systems - from attacking imitation learning policies \cite{jia2022physical} to deceiving reinforcement learning agents \cite{bai2025rat} - these methods were designed for relatively simple modular architectures. Previous work has shown that adversarial patches \cite{brown2018adversarialpatch, 11094245} and objects \cite{tsai2020robust} can fool perception systems in autonomous vehicles \cite{deng2020analysis}, yet these approaches cannot be directly transferred to VLA models. The fundamental difference lies in VLA's architectural complexity: Their multimodal fusion mechanisms, diffusion-based generation processes, and temporal decision-making create novel vulnerability patterns that require entirely new attack strategies to understand and ultimately defend against.

In this work, we present the first systematic analysis of mainstream VLA models' vulnerabilities to physical adversarial attacks, bridging the gap between theoretical security concerns and practical threats. We propose a Universal Adversarial Object (UAO)—a seemingly innocuous 3cm textured sphere—that, when strategically placed within the robot's workspace, severely disrupts VLA model decision-making across diverse tasks and viewing conditions. As illustrated in Figure \ref{fig1}, this simple object induces dramatic behavioral corruptions: robots exhibit severe trajectory deviations, erratic oscillations, and catastrophic task failures, with success rates plummeting by up to 70\% in precision manipulation tasks. Our multi-level attack framework simultaneously targets trajectory planning, task execution, and motion control, exploiting the 
\newpage  
\setlength{\topmargin}{-0.25in}  
\setlength{\textheight}{9.25in}  
\noindent  
hierarchical nature of VLA processing to maximize disruption while maintaining physical realizability and visual inconspicuousness.

Specifically, we design a multi-level attack framework that disrupts the entire perception-to-action decision chain of VLA models: trajectory planning, task execution, and action control. 
Given that VLA models are based on diffusion architectures\cite{peebles2023scalable} with inaccessible internal gradients, we adopt a black-box optimization strategy. 
To ensure the attack's sustained efficacy, we integrate temporal characteristics and the SPSA optimization method to enhance efficiency. Additionally, we employ data augmentation techniques to guarantee the robustness and physical feasibility of the adversarial objects under various environmental conditions.

We thoroughly evaluate our approach on two representative VLA models (Pi0 and RDT) across 13 robotic manipulation tasks with varying complexities. Experimental results show that, in simulation, the adversarial object reduces task success rates by 31.2\%-39.9\%, and in more complex environments, the success rate drops to nearly zero (RDT: 2.1\%, Pi0: 1.9\%). Moreover, we demonstrate UAO on a real dual-arm robotic platform, achieving a sim-to-real transfer rate of 81.4\%-82.2\%. Even when the UAO is visible in only a single camera view, it leads to a considerable decline, highlighting its viewpoint robustness. 

The main contributions of this work include:

\textbf{(1) Comprehensive revelation of VLA model adversarial vulnerability.} We show through large-scale experiments that VLA models are highly vulnerable to adversarial objects, which can severely disrupt decision-making.

\textbf{(2) Multi-level attack framework and UAO generation.} We introduce a multi-level attack framework that generates robust adversarial objects, overcoming the limitations of traditional attack methods.

\textbf{(3) Validation from simulation to real-world deployment.} We validate the attack effectiveness in both simulated and real-world environments, demonstrating its practical threat to VLA model safety.

\section{Related Works}
\label{related}

\noindent\textbf{Vision-Language-Action Models.} 

Vision-Language-Action (VLA) models represent a paradigm shift in robotic learning, unifying visual perception, language understanding, and action generation within end-to-end frameworks \cite{ma2024survey, sapkota2025vision}. Early VLA explorations primarily constructed hierarchical frameworks leveraging the perceptual and reasoning capabilities of Vision-Language Models (VLMs) \cite{hu2024look, li2024manipllm, huang2024manipvqa, jin2024robotgpt}. Subsequent advances encoded actions as an additional modality, achieving superior generalization through joint VLA training and enabling end-to-end learning for complex manipulation tasks \cite{jiang2022vima, brohan2022rt, zitkovich2023rt, kim2024openvla, wen2025dexvla}.

Recent VLA models have achieved significant breakthroughs in architectural design and training strategies \cite{sapkota2025vision}. In particular, Pi0 \cite{black2024pi} introduces innovative policy learning methods that unify various manipulation tasks through large-scale demonstration data. RDT \cite{liurdt} incorporates diffusion models for action generation, producing more precise trajectories through iterative denoising. Although these models demonstrate impressive performance on multiple benchmarks \cite{mu2024robotwin}, existing research focuses primarily on improving capabilities, overlooking systematic security analysis \cite{liu2025survey}.

\noindent\textbf{Adversarial Attacks in Robotic.} 

Adversarial attacks, which mislead deep learning models through carefully crafted input perturbations, have emerged as critical tools to evaluate and improve model robustness \cite{zhang2021survey}. The progression from digital to physical-world attacks has exposed the pervasive vulnerability of deep models to adversarial perturbations. Physical adversarial attacks have proven particularly significant, and methods such as adversarial patches \cite{brown2018adversarialpatch, 11094245} and adversarial objects \cite{tsai2020robust} demonstrate the feasibility of deceiving vision models in real-world environments. These attacks pose serious security threats across domains including autonomous driving \cite{deng2020analysis}, facial recognition \cite{liu2025projattacker}, and object detection \cite{wei2024revisiting}.

Robotic systems present unique security challenges, requiring continuous decision-making and physical interaction in dynamic environments. Early adversarial research primarily targeted traditional small-scale robotic models, such as imitation learning \cite{jia2022physical} and reinforcement learning policies \cite{bai2025rat}. However, the emergence of VLA models marks a fundamental shift toward large-scale, multimodal, end-to-end architectures, introducing novel attack surfaces and security challenges. The complex multimodal interactions and diffusion-based architectures of VLA models render traditional attack methods ineffective. Although Wang et al. \cite{wang2025exploringadversarialvulnerabilitiesvisionlanguageaction} pioneered the exploration of vulnerabilities in VLA, their analysis remained limited to specific models and tasks. In contrast, we propose the first universal physical adversarial attack framework that systematically compromises VLA models across different architectures and validates its effectiveness on real robotic platforms.

\section{Methodology}
\label{sec:method}

\subsection{Preliminary}

Currently, VLA models leverage end-to-end learning, directly transforming multimodal inputs into robot control commands \cite{ma2024survey, sapkota2025vision}.
Specifically, a VLA model $\mathcal{M}$ receives several types of inputs at each decision timestep $t$. These include task instructions $\mathcal{L}$ in natural language describing the objectives; visual observations $\{\mathbf{I}_{i,t}\}_{i=1}^{N_{cam}}$ from multiple viewpoints; and the robot's proprioceptive state $\mathbf{s}_t \in \mathcal{S}$, encompassing joint angles and end-effector positions.

The output of the VLA model comprises two components: a predicted action sequence $\mathbf{a}_{t:t+k} = \{\mathbf{a}_t, \mathbf{a}_{t+1}, ..., \mathbf{a}_{t+k}\}$ where $k$ denotes the prediction horizon and a task completion indicator $r \in \{0, 1\}$. In this work, we focus on models operating on a 7-DoF dual-arm Aloha platform \cite{fu2024mobile}. At each timestep, actions are represented as:

\[
\mathbf{a}_t = [\Delta P_x, \Delta P_y, \Delta P_z, \Delta R_x, \Delta R_y, \Delta R_z, g]
\]

where $\Delta P_x, \Delta P_y, \Delta P_z$ and $\Delta R_x, \Delta R_y, \Delta R_z$ denote relative position and rotation changes along the $x$, $y$, and $z$ axes respectively, and $g$ is a binary gripper state.

\subsection{Problem Definition}

We formalize the adversarial attack as finding a universal visual perturbation that induces task failure without modifying model parameters $\mathcal{M}$, language instructions $\mathcal{L}$, or robot states $\mathcal{S}$. Specifically, we design an adversarial patch $\mathbf{P} \in \mathbb{R}^{H \times W \times 3}$, where $H$ and $W$ define spatial resolution. This patch is projected onto a sphere of radius $r$ via a spherical mapping function $\phi: \mathbb{R}^{H \times W \times 3} \rightarrow \mathbb{S}^2(r)$, forming a universal adversarial object $\mathbf{O_p}$. When placed in position $\mathbf{c} \in \mathbb{R}^3$ within the robot workspace, this object appears in camera views $\mathbf{I}_{i,t}$ and disrupts task execution through visual manipulation.

We employ black-box optimization to generate this universal patch, accessing only model outputs without internal parameters or gradients. Our objective is to find an optimal patch $\mathbf{P}^*$ such that the resulting trajectory $\tau^{adv} = \{\mathbf{a}_1^{adv}, ..., \mathbf{a}_T^{adv}\}$ significantly deviates from the clean trajectory $\tau^{clean}$ and induces task failure ($r^{adv} = 0$). To achieve this, we design a multi-objective loss function incorporating trajectory deviation, task failure, action perturbation, and visual naturalness.

\subsection{Multi-level Attack Framework for VLA}

\begin{figure*}[ht!]
    \centering
    \includegraphics[width=1.0\textwidth]{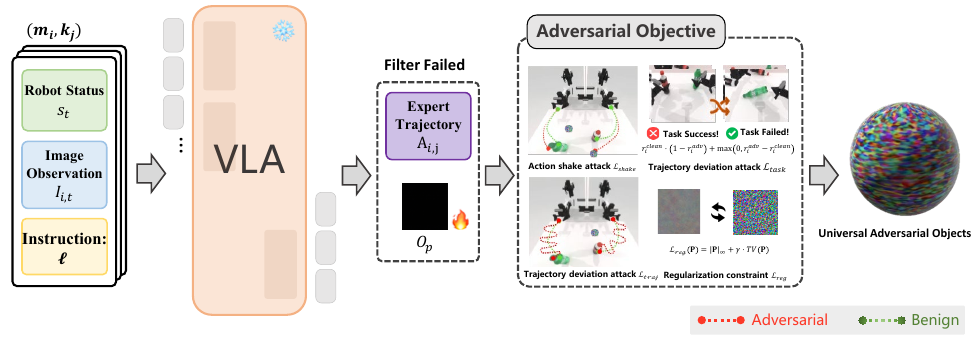}
    \caption{
    The framework for generating universal adversarial objects via multi-level attack optimization. Starting from expert demonstrations, we filter successful trajectories and optimize adversarial textures using four complementary losses that target different aspects of robot behavior—from high-level task failure to low-level motion perturbations. The resulting texture pattern is mapped onto a sphere to create a physically universal adversarial object.
    }
    \label{method}
\end{figure*}

The generation of VLA actions involves multiple processing levels, from pixel-level visual perception to task-level goal understanding and trajectory-level action planning \cite{kim2024openvla, din2025vision}. Based on this observation, we design a hierarchical attack framework that jointly disrupts these levels for comprehensive adversarial impact.

\subsubsection{Trajectory Deviation Attack}

Robotic manipulation tasks exhibit strong temporal dependencies, with VLA models demonstrating distinct phases during long-horizon execution \cite{takano2021continuous, goldmanevaluating}: coarse positioning in initial stages, primary manipulation in intermediate phases, and precise control for task completion. We design a time-weighted trajectory deviation loss:

\[
\mathcal{L}_{traj}(\mathbf{P}) = \frac{1}{T} \sum_{t=1}^{T} w_t \cdot \|\tau_{clean}^t - \tau_{adv}^t\|_2
\]

where $\tau_{clean}^t$ and $\tau_{adv}^t$ denote end-effector positions at time $t$ under clean and adversarial conditions, respectively, with $\|\cdot\|_2$ computing Euclidean distance. The temporal weight function $w_t = \exp(\alpha \cdot t/T)$ assigns exponentially increasing importance to later timesteps, concentrating attack energy on critical decision moments. Parameter $\alpha$ controls the growth rate, with larger values emphasizing late-stage precision control.

\subsubsection{Task Failure Attack}

The success rate is the primary metric for robotic manipulation, yet its discrete binary nature prevents direct gradient optimization. Moreover, diffusion-based VLA models preclude direct access to internal reward signals. To address this challenge, we propose a contrastive-inspired task failure loss \cite{koch2015siamese}:

\[
\mathcal{L}_{task}(\mathbf{P}) = \sum_{i=1}^{N} r_i^{clean} \cdot (1 - r_i^{adv}) + \beta \cdot \max(0, r_i^{adv} - r_i^{clean})
\]

The first term targets originally successful tasks, generating negative loss when they fail under adversarial conditions. The second term penalizes unintended improvements where failed tasks become successful. The asymmetric penalty coefficient $\beta > 1$ reflects the attack's directional nature.

\subsubsection{Action Perturbation Attack}

Through imitation learning, VLA models acquire smooth action primitives by mimicking expert demonstrations \cite{kim2024openvla, liurdt}. We exploit this by inducing high-frequency directional changes that disrupt the diffusion denoising process. Our action perturbation loss is:

\[
\mathcal{L}_{shake}(\mathbf{P}) = -\frac{1}{T-2}\sum_{t=2}^{T-1} \|\mathbf{v}_{t+1} - \mathbf{v}_t\|_2
\]

where $\mathbf{v}_t = (\tau_t^{adv} - \tau_{t-1}^{adv})/\|\tau_t^{adv} - \tau_{t-1}^{adv}\|_2$ represents the normalized direction of action. This loss maximizes the directional differences between adjacent timesteps, generating oscillatory patterns. Given velocity constraints in robotic systems, we perturb direction rather than magnitude. Geometrically, $\|\mathbf{v}_{t+1} - \mathbf{v}_t\|_2 \in [0, 2]$, where 0 indicates a linear action and 2 represents a 180-degree reversal.

\subsubsection{Regularization Constraints}

To ensure physical realizability and inconspicuousness, we introduce regularization constraints \cite{evtimov2017robust, eykholt2018robust}:

\[
\mathcal{L}_{reg}(\mathbf{P}) = \|\mathbf{P}\|_\infty + \gamma \cdot TV(\mathbf{P})
\]

The infinity norm $\|\mathbf{P}\|_\infty$ limits the maximum pixel intensity within the standard RGB range. Regularization of total variation $TV(\mathbf{P}) = \sum_{i,j} \sqrt{(P_{i+1,j} - P_{i,j})^2 + (P_{i,j+1} - P_{i,j})^2}$ promotes spatial smoothness by penalizing adjacent pixel differences. This smoothness yields multiple benefits: natural texture appearance that reduces detection risk, robustness to printing errors and lighting variations, and prevention of high-frequency local optima.

\subsubsection{Overall Optimization Objective}

In summary, our total loss function is defined as follows:

\[
\begin{split}
\mathcal{L}_{total}(\mathbf{P}) = \lambda_1 \mathcal{L}_{traj}(\mathbf{P}) + \lambda_2 \mathcal{L}_{task}(\mathbf{P}) \\
+ \lambda_3 \mathcal{L}_{shake}(\mathbf{P}) + \lambda_4 \mathcal{L}_{reg}(\mathbf{P})
\end{split}
\]

The weight coefficients reflect relative importance, with empirically determined values: $\lambda_1 = 0.30$, $\lambda_2 = 0.45$, $\lambda_3 = 0.20$, $\lambda_4 = 0.05$. Task failure receives the highest weight as the primary objective, whereas trajectory deviation and action perturbation serve as auxiliary targets. Regularization maintains minimal weight for necessary physical constraints.

\subsection{Optimization Strategy}

Considering the training cost, we develop an efficient black-box optimization method achieving effective attacks within limited query budgets.

\subsubsection{Gradient Estimation}

Given the inaccessibility of internal gradients in VLA models, we adopt Simultaneous Perturbation Stochastic Approximation (SPSA) \cite{spall2002multivariate}, which efficiently estimates high-dimensional gradients through merely two function evaluations:

\[
\nabla_{\mathbf{P}} \mathcal{L} \approx \frac{\mathcal{L}(\mathbf{P} + c\Delta) - \mathcal{L}(\mathbf{P} - c\Delta)}{2c} \cdot \Delta^{-1}
\]

where the perturbation vector $\Delta \in \{-1, +1\}^{H \times W \times 3}$ has elements sampled independently of the Bernoulli distribution. Step size $c$ follows an adaptive schedule: larger values for initial exploration, gradually decreasing for fine-tuning.

\subsubsection{Data Augmentation}

During manipulation, camera viewpoints continuously change as the arm moves, altering the patch's appearance and potentially occluding regions. To ensure attack effectiveness under dynamic conditions, we apply augmentations during optimization:

\[
\mathbf{P}_{aug} = \mathcal{T}_{rot}(\theta) \circ \mathcal{T}_{scale}(s) \circ \mathcal{T}_{shift}(\delta) (\mathbf{P})
\]

where the rotation angle $\theta \sim \mathcal{U}(0, 2\pi)$ ensures orientation invariance, the scale factor $s \sim \mathcal{U}(0.8, 1.2)$ handles distance variations, and the cyclic shift $\delta$ simulates the rotation of the sphere. Color enhancement adjusts contrast $\alpha \sim \mathcal{U}(0.8, 1.2)$ and brightness $\beta \sim \mathcal{U}(-0.1, 0.1)$ to provide light stability.

\subsubsection{Optimization Process}

We employ alternating optimization across models and tasks to ensure broad attack effectiveness. For each model-task pair $(\mathbf{m},\mathbf{k})$, we evaluate clean performance as baseline, then use SPSA to estimate gradients and update the patch—requiring only two evaluations per iteration. Tasks that initially fail are filtered out. The optimization is implemented in RoboTwin simulator \cite{mu2024robotwin} with PyOpenGL\footnote{\url{https://github.com/mcfletch/pyopengl}} for sphere texture mapping. Algorithm \ref{alg:patch_optimization} details the complete process, generating universal patches effective in multiple VLA models and tasks.

\begin{algorithm}
\caption{Universal Adversarial Object Optimization Process.}
\label{alg:patch_optimization}
\begin{algorithmic}[1]
\STATE {\bfseries Input:} Model set $\mathcal{M}$, task set $\mathcal{K}$, iterations $N$
\STATE {\bfseries Initialize:} $\mathbf{P} \sim \mathcal{U}(-0.01, 0.01)^{H \times W \times 3}$
\FOR{$n = 1, 2, \cdots, N$}
\FOR{$(m, k) \in \mathcal{M} \times \mathcal{K}$}
\STATE $r^{clean}, \tau^{clean} \leftarrow \text{Evaluate}(m, k, \emptyset)$
\IF{$r^{clean} = 0$}
\STATE \textbf{continue}
\ENDIF
\STATE $\mathbf{P}_{aug} \leftarrow \mathcal{T}(\mathbf{P})$ \COMMENT{Augmentation}
\STATE $\Delta \sim \{-1, +1\}^{H \times W \times 3}$ 
\STATE $\mathcal{L}^{\pm} \leftarrow \mathcal{L}_{total}(\text{Evaluate}(m, k, \mathbf{P}_{aug} \pm c\Delta))$
\STATE $\mathbf{P} \leftarrow \mathbf{P} - \eta \cdot \frac{\mathcal{L}^+ - \mathcal{L}^-}{2c} \cdot \Delta^{-1}$ \COMMENT{SPSA update}
\ENDFOR
\ENDFOR
\STATE {\bfseries Output:} Optimized patch $\mathbf{P}^*$
\end{algorithmic}
\end{algorithm}

\section{Experiment and Analysis}
\label{sec:experiment}

This section verifies the proposed UAO attack method through systematic experiments. Section \ref{sec:exp_setup} first introduces the experimental setup, and Section \ref{sec:main_results} presents the main attack results. Subsequently, Section \ref{sec:ablation} analyzes the contribution of each loss component through ablation experiments, Section \ref{sec:patch_analysis} explores the impact of key hyperparameters via parameter analysis, Section \ref{sec:multi_view} tests the view robustness through multi-view evaluation, and finally, Section \ref{sec:real_world} conducts real-world validation.

\begin{table*}[ht!]
\centering
\caption{Attack performance on different VLA models and tasks. SR: Success Rate, $\Delta$SR: Success Rate Drop, TD: Trajectory Deviation. All metrics are averaged over 10 independent runs.}
\label{tab:main_results}
\resizebox{\textwidth}{!}{%
\begin{tabular}{l|cccc|cccc|cccc|cccc}
\toprule
\multirow{3}{*}{\textbf{Task}} & \multicolumn{8}{c|}{\textbf{RDT}} & \multicolumn{8}{c}{\textbf{Pi0}} \\
\cmidrule{2-17}
& \multicolumn{4}{c|}{\textbf{Easy}} & \multicolumn{4}{c|}{\textbf{Hard}} & \multicolumn{4}{c|}{\textbf{Easy}} & \multicolumn{4}{c}{\textbf{Hard}} \\
\cmidrule{2-5} \cmidrule{6-9} \cmidrule{10-13} \cmidrule{14-17}
& \textbf{Clean SR} & \textbf{Adv. SR} & \textbf{$\Delta$SR↓} & \textbf{TD} & \textbf{Clean SR} & \textbf{Adv. SR} & \textbf{$\Delta$SR↓} & \textbf{TD} & \textbf{Clean SR} & \textbf{Adv. SR} & \textbf{$\Delta$SR↓} & \textbf{TD} & \textbf{Clean SR} & \textbf{Adv. SR} & \textbf{$\Delta$SR↓} & \textbf{TD} \\
\midrule
Adjust Bottle           & 83\% & 38\% & 45\% & 0.24 & 73\% & 5\%  & 68\% & 0.29 & 88\% & 44\% & 44\% & 0.26 & 58\% & 2\%  & 56\% & 0.31 \\
Beat Block Hammer       & 75\% & 35\% & 40\% & 0.22 & 39\% & 3\%  & 36\% & 0.27 & 45\% & 22\% & 23\% & 0.18 & 19\% & 0\%  & 19\% & 0.25 \\
Click Alarmclock        & 62\% & 30\% & 32\% & 0.19 & 10\% & 0\%  & 10\% & 0.21 & 61\% & 28\% & 33\% & 0.20 & 13\% & 0\%  & 13\% & 0.23 \\
Grab Roller             & 72\% & 33\% & 39\% & 0.21 & 45\% & 4\%  & 41\% & 0.26 & 94\% & 46\% & 48\% & 0.27 & 78\% & 8\%  & 70\% & 0.32 \\
Dump Bin Bigbin         & 66\% & 28\% & 38\% & 0.20 & 30\% & 2\%  & 28\% & 0.25 & 81\% & 35\% & 46\% & 0.26 & 26\% & 1\%  & 25\% & 0.28 \\
Open Laptop             & 57\% & 25\% & 32\% & 0.19 & 34\% & 3\%  & 31\% & 0.24 & 87\% & 40\% & 47\% & 0.27 & 44\% & 2\%  & 42\% & 0.29 \\
Press Stapler           & 43\% & 21\% & 22\% & 0.15 & 22\% & 1\%  & 21\% & 0.22 & 60\% & 32\% & 28\% & 0.18 & 31\% & 2\%  & 29\% & 0.26 \\
Put Object Cabinet      & 31\% & 15\% & 16\% & 0.13 & 20\% & 1\%  & 19\% & 0.21 & 70\% & 35\% & 35\% & 0.21 & 16\% & 0\%  & 16\% & 0.24 \\
Shake Bottle Horiz.     & 82\% & 37\% & 45\% & 0.25 & 53\% & 3\%  & 50\% & 0.29 & 97\% & 48\% & 49\% & 0.28 & 53\% & 2\%  & 51\% & 0.31 \\
Shake Bottle            & 76\% & 36\% & 40\% & 0.22 & 43\% & 2\%  & 41\% & 0.27 & 95\% & 45\% & 50\% & 0.28 & 62\% & 4\%  & 58\% & 0.30 \\
Pick Dual Bottles       & 40\% & 18\% & 22\% & 0.15 & 15\% & 0\%  & 15\% & 0.20 & 59\% & 28\% & 31\% & 0.19 & 10\% & 0\%  & 10\% & 0.18 \\
Place Burger Fries      & 52\% & 23\% & 29\% & 0.17 & 25\% & 1\%  & 24\% & 0.23 & 78\% & 36\% & 42\% & 0.24 & 6\%  & 0\%  & 6\%  & 0.15 \\
Open Microwave          & 35\% & 15\% & 20\% & 0.14 & 22\% & 1\%  & 21\% & 0.22 & 82\% & 39\% & 43\% & 0.25 & 48\% & 3\%  & 45\% & 0.28 \\
\midrule
\textbf{Average} & 57.1\% & 25.7\% & \textbf{32.3\%} & \textbf{0.19} 
                 & 33.5\% & 2.1\% & \textbf{31.2\%} & \textbf{0.25} 
                 & 75.1\% & 36.4\% & \textbf{39.9\%} & \textbf{0.24} 
                 & 35.4\% & 1.9\% & \textbf{33.8\%} & \textbf{0.27} \\
\bottomrule
\end{tabular}
}
\end{table*}

\subsection{Experiment Setup}
\label{sec:exp_setup}

\subsubsection{Target Models}

We select Pi0 \cite{black2024pi} and RDT \cite{liurdt} as the target VLA models. Pi0 uses a Transformer-based architecture, with the pre-trained PaLI-Gemma \cite{team2024gemma} backbone, and generates continuous action trajectories using Flow Matching. RDT, with a Diffusion Transformer architecture \cite{peebles2023scalable} and 1.2 billion parameters, is pre-trained on 46 multi-robot datasets (over 1 million trajectories) and fine-tuned with 6000 dual-arm operation data. Both models are extensively pre-trained and fine-tuned, showcasing excellent generalization in dual-arm manipulation tasks.

\subsubsection{Tasks and Environment}

We use the RoboTwin simulator as the training and evaluation platform, designed specifically for dual-arm coordination, and fully supports Pi0 and RDT models. The simulator includes a benchmark with expert trajectories. From 50 tasks, we select 13 representative ones with high success rates, ranging from basic grasping to assembly and dual-arm coordination tasks. Each task has an Easy and Hard mode—Easy uses a clean desktop, while Hard introduces random lighting, material, and irrelevant objects. The experiments use the standard Aloha dual-arm configuration, with 7 degrees of freedom per arm and three fixed cameras (one overhead, two at the end) for multi-view observation. The UAO $\mathbf{O}_p$ is generated by mapping optimized patch $\mathbf{P}$ onto the surface of a 3cm radius sphere, randomly placed in the workspace during each evaluation, ensuring visibility in at least two camera views.

\subsubsection{Evaluation Metrics}

We use two complementary metrics to evaluate the attack's effect. The first is task success rate, which measures the proportion of successfully completed tasks in both clean and adversarial environments, reflecting the overall impact of the attack. The second is Trajectory Discrepancy (TD), which quantifies deviations in the robot's action:

\[
TD = \frac{1}{T}\sum_{t=1}^{T} \frac{\|\tau_{clean}^t - \tau_{adv}^t\|_2}{\|\tau_{clean}^{T} - \tau_{clean}^{0}\|_2}
\]

where $\tau_{clean}^t$ and $\tau_{adv}^t$ represent the end-effector positions at time $t$ in the clean and adversarial environments. The denominator normalizes by the total trajectory length, making deviations comparable between tasks. TD quantifies the impact of the perturbation even if the task is not completely failed. Each task is evaluated with 10 different random seeds, and the mean result is reported.

\subsubsection{Implementation Details}

\begin{figure}[!t]
    \centering
    \includegraphics[width=0.35\textwidth]{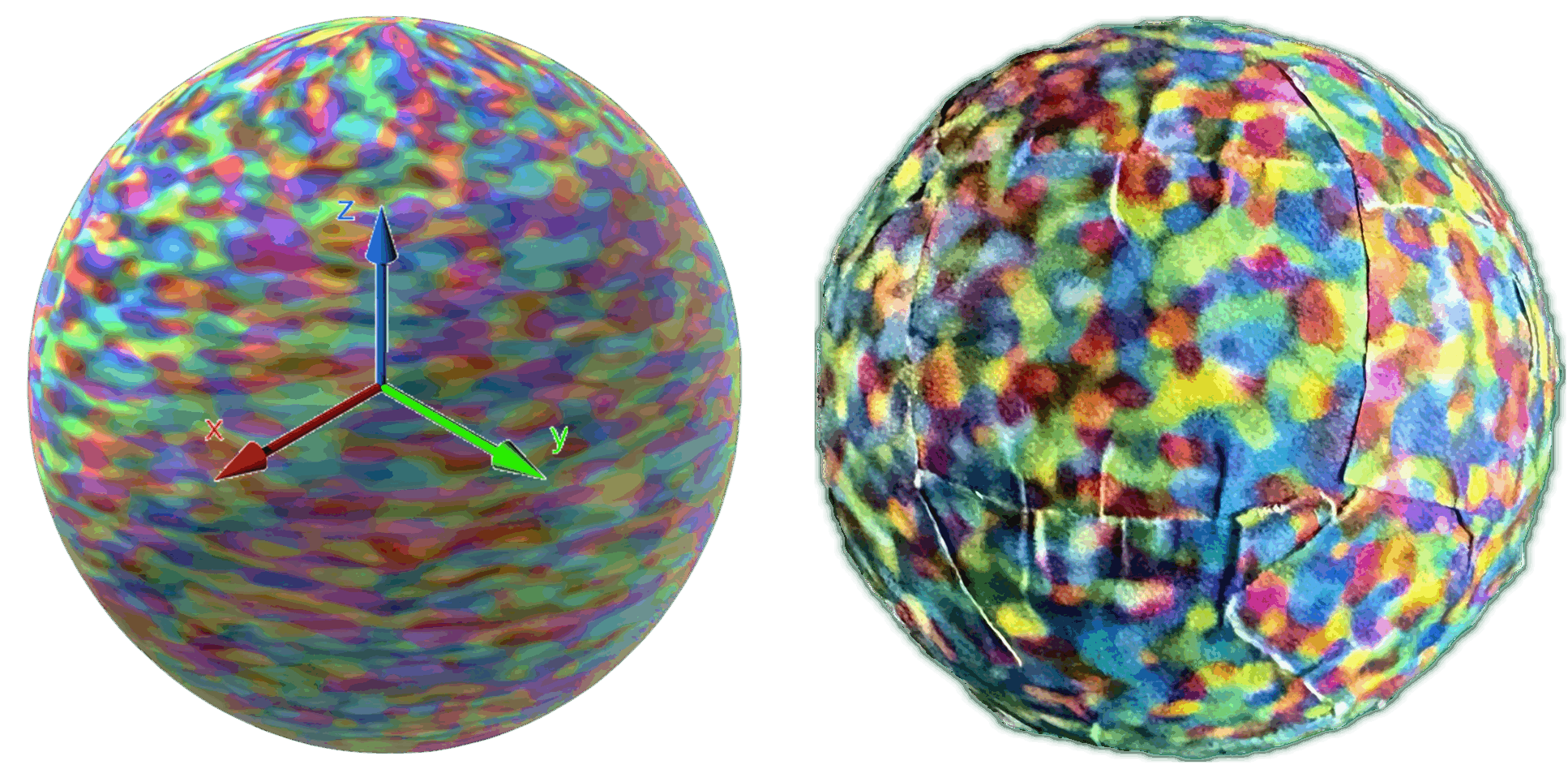}
    \caption{The image of universal adversarial objects in simulator and real-world.}
    \label{object}
\end{figure}

The adversarial patch $\mathbf{P}$ is set at a resolution of $128 \times 128$ pixels and projected onto the surface of a 3 cm radius sphere using spherical mapping. The simulation and real-world objects are shown in Figure \ref{object}. The optimization process uses the SPSA algorithm with an initial perturbation step size of $c = 0.01$, decaying at a rate of 0.95, and a learning rate of $\eta = 0.001$. For each model-task pair, we train for 500 iterations, with optimization taking approximately one week on an NVIDIA A100 80G GPU. The optimized patch remains fixed during the testing phase.

\subsection{Main Results}
\label{sec:main_results}

Table \ref{tab:main_results} presents comprehensive evaluation results of our UAO against the RDT and Pi0 models in 13 representative manipulation tasks. The experiments reveal pervasive vulnerabilities in state-of-the-art VLA models, with a single 3cm textured sphere inducing average success rate drops of 31.2\%-39.9\% across all scenarios. Beyond binary task failure, the trajectory deviation metric quantifies systematic behavioral corruption, demonstrating 19\%-27\% positional errors throughout execution sequences even in partially successful attempts. This dual measurement reveals that adversarial perturbations not only prevent task completion but fundamentally distort the learned visuomotor mappings underlying robotic control.

The data exposes a critical vulnerability gradient correlated with environmental complexity and task precision requirements. In Hard mode, where visual distractors and lighting variations are introduced, both models catastrophically fail with near-zero success rates (RDT: 2.1\%, Pi0: 1.9\%), compared to 25.7\%-36.4\% residual performance in Easy mode. This dramatic amplification suggests that VLA models develop increased dependence on visual features when navigating complex scenes, inadvertently expanding their attack surface. Task-specific analysis reveals heightened susceptibility in precision-demanding operations: Grab Roller (70\% drop for Pi0-Hard), Adjust Bottle (68\% drop for RDT-Hard), and dual-arm coordination tasks consistently exhibit vulnerability rates exceeding 50\%. Conversely, tasks with larger error tolerance such as Press Stapler and Put Object Cabinet demonstrate relative resilience, though still experiencing substantial degradation (16\%-29\% drops). Notably, Pi0 exhibits marginally higher baseline performance but suffers greater absolute degradation under attack, while RDT's diffusion-based architecture shows no inherent robustness advantage despite its iterative refinement mechanism. These findings underscore a fundamental security gap in current VLA architectures: the same multimodal integration that enables impressive generalization simultaneously creates exploitable dependencies that can be systematically compromised through carefully crafted visual perturbations.

\subsection{Ablation Study}
\label{sec:ablation}

\begin{table}[!t]
\centering
\caption{Ablation study on different attack components. w/o indicates that the component is removed.}
\label{tab:ablation}
\resizebox{\columnwidth}{!}{%
\begin{tabular}{l|cc|cc|cc|cc}
\toprule
\multirow{2}{*}{\textbf{Method}} & \multicolumn{2}{c|}{\textbf{RDT (Easy)}} & \multicolumn{2}{c|}{\textbf{RDT (Hard)}} & \multicolumn{2}{c|}{\textbf{Pi0 (Easy)}} & \multicolumn{2}{c}{\textbf{Pi0 (Hard)}} \\
\cmidrule{2-9}
& \textbf{$\Delta$SR↓} & \textbf{TD} & \textbf{$\Delta$SR↓} & \textbf{TD} & \textbf{$\Delta$SR↓} & \textbf{TD} & \textbf{$\Delta$SR↓} & \textbf{TD} \\
\midrule
Full Attack & \textbf{32.3}\% & 0.19 & \textbf{31.2}\% & 0.25 & \textbf{39.9}\% & 0.24 & \textbf{33.8}\% & 0.27 \\
\midrule
w/o $\mathcal{L}_{traj}$ & 22.9\% & 0.12 & 24.8\% & 0.16 & 26.8\% & 0.15 & 26.2\% & 0.18 \\
w/o $\mathcal{L}_{task}$ & 18.6\% & 0.18 & 18.2\% & 0.23 & 22.4\% & 0.22 & 18.6\% & 0.25 \\
w/o $\mathcal{L}_{shake}$ & 25.7\% & 0.14 & 27.9\% & 0.19 & 32.9\% & 0.18 & 30.1\% & 0.21 \\
w/o $\mathcal{L}_{reg}$ & 32.3\% & 0.20 & 31.2\% & 0.26 & 39.2\% & 0.25 & 33.7\% & 0.28 \\
\midrule
Only $\mathcal{L}_{task}$ & 14.8\% & 0.08 & 14.9\% & 0.11 & 18.7\% & 0.10 & 15.3\% & 0.13 \\
Only $\mathcal{L}_{traj}$ & 11.4\% & 0.16 & 11.1\% & 0.21 & 15.9\% & 0.19 & 11.7\% & 0.23 \\
Only $\mathcal{L}_{shake}$ & 8.2\% & 0.11 & 7.7\% & 0.15 & 11.6\% & 0.14 & 8.1\% & 0.17 \\
\bottomrule
\end{tabular}
}
\end{table}

To systematically assess the contributions of each component in our multi-level attack framework, we conducted comprehensive ablation experiments by selectively removing individual loss terms. The results in Table \ref{tab:ablation} reveal a hierarchical importance structure among the components. The task failure loss $\mathcal{L}_{task}$ emerges as the most critical element, with its removal causing the success rate drop to plummet from 32.3\% to 18.6\%, underscoring its fundamental role in directly optimizing the primary attack objective. The trajectory deviation loss $\mathcal{L}_{traj}$ demonstrates substantial impact as well, contributing approximately 9.4 percentage points to the overall attack effectiveness by inducing systematic positional errors throughout the manipulation sequence. The action perturbation loss $\mathcal{L}_{shake}$ proves essential for disrupting the smooth action primitives learned through imitation, with its absence reducing attack potency by 6.6\%. While the regularization term $\mathcal{L}_{reg}$ exhibits minimal quantitative impact on success rates, it serves a crucial qualitative role in maintaining physical realizability and visual inconspicuousness, preventing the optimization from converging to easily detectable high-frequency patterns. Most notably, when restricted to single loss components, the attack effectiveness catastrophically degrades ($\Delta$SR merely 8.2\%-18.7\%), demonstrating that the synergistic interaction between multiple attack vectors is fundamental to compromising VLA models' robust decision-making pipeline.

\begin{table}[!t]
\centering
\caption{Attack performance under different camera visibility conditions. The adversarial object is placed to be visible in different numbers of cameras.}
\label{tab:multi_view}
\resizebox{\columnwidth}{!}{%
\begin{tabular}{l|cc|cc|cc|cc}
\toprule
\multirow{2}{*}{\textbf{Visibility}} & \multicolumn{2}{c|}{\textbf{RDT (Easy)}} & \multicolumn{2}{c|}{\textbf{RDT (Hard)}} & \multicolumn{2}{c|}{\textbf{Pi0 (Easy)}} & \multicolumn{2}{c}{\textbf{Pi0 (Hard)}} \\
\cmidrule{2-9}
& \textbf{$\Delta$SR↓} & \textbf{TD} & \textbf{$\Delta$SR↓} & \textbf{TD} & \textbf{$\Delta$SR↓} & \textbf{TD} & \textbf{$\Delta$SR↓} & \textbf{TD} \\
\midrule
3 Cameras (All) & \textbf{32.3}\% & 0.19 & \textbf{31.2}\% & 0.25 & \textbf{39.9}\% & 0.24 & \textbf{33.8}\% & 0.27 \\
2 Cameras & 26.1\% & 0.15 & 24.7\% & 0.20 & 31.2\% & 0.19 & 26.9\% & 0.22 \\
1 Camera & 15.8\% & 0.09 & 14.3\% & 0.12 & 18.6\% & 0.11 & 15.2\% & 0.13 \\
\midrule
Top Camera Only & 12.4\% & 0.08 & 11.8\% & 0.11 & 14.9\% & 0.10 & 12.7\% & 0.12 \\
Wrist Cameras Only & 17.9\% & 0.10 & 16.2\% & 0.13 & 21.3\% & 0.12 & 17.4\% & 0.14 \\
\bottomrule
\end{tabular}
}
\end{table}

\begin{figure*}[!t]
    \centering
    \includegraphics[width=1.0\textwidth]{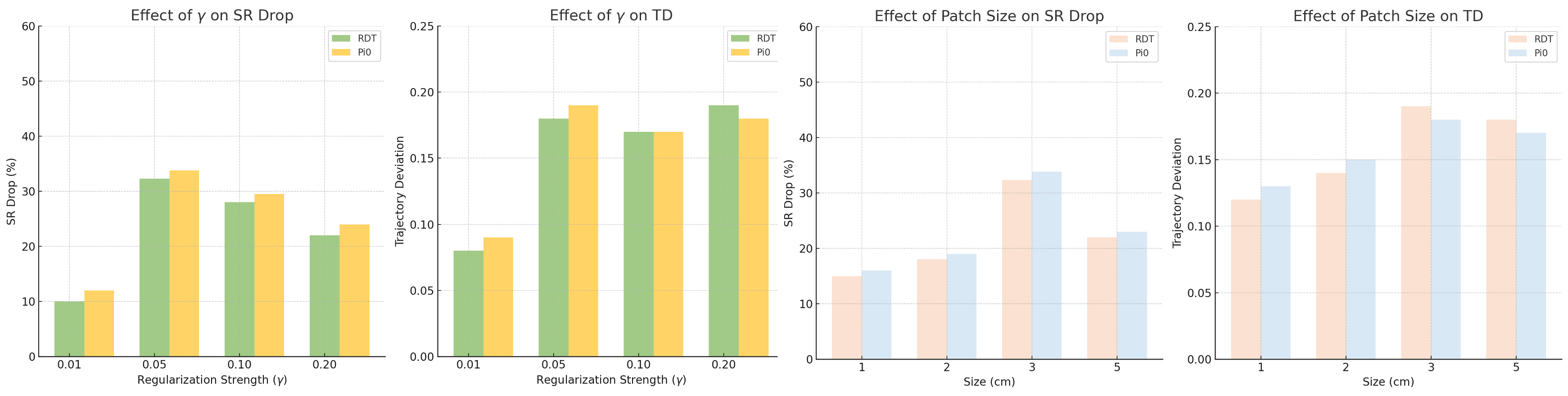}
    \caption{Effect of Patch Size and Regularization Strength on SR Drop and Trajectory Deviation. Moderate patch sizes (3cm) maximize attack impact while increased regularization ($\gamma > 0.1$) significantly diminishes success rate reduction but paradoxically increases trajectory perturbation.}
    \label{fig:patch_analysis}
\end{figure*}

\subsection{Patch Parameter Analysis}
\label{sec:patch_analysis}

To understand the sensitivity of our attack to key hyperparameters, we systematically investigated the influence of patch size and regularization strength on both attack effectiveness and trajectory perturbation, as illustrated in Figure \ref{fig:patch_analysis}. The relationship between patch size and attack potency exhibits a distinctive nonmonotonic pattern: moderate patches (3cm radius) achieve optimal performance by maintaining sufficient visual saliency while remaining within the robots' operational field of view throughout task execution. Smaller patches (1-2cm) suffer from reduced visual impact due to limited pixel coverage in the captured images, while larger patches (5cm) paradoxically weaken the attack by frequently exceeding camera boundaries during dynamic manipulation, resulting in intermittent visibility. The regularization parameter $\gamma$ reveals an intriguing trade-off between stealth and disruption—increased regularization strength enhances visual naturalness through smoother texture transitions, yet counterintuitively amplifies trajectory deviation from 0.08 to 0.19, suggesting that constrained perturbations force the optimization to exploit more subtle but kinematically disruptive features. This phenomenon indicates that VLA models may be particularly vulnerable to smooth, naturalistic perturbations that align with expected environmental textures while fundamentally corrupting the learned visuomotor mappings.

\subsection{Multi-view Robustness Evaluation}
\label{sec:multi_view}

In real-world robotic operations, the visibility of objects in cameras changes with the arm's movement. Table \ref{tab:multi_view} evaluates the effectiveness of the attack under different viewpoint conditions. Even when the adversarial object is visible in only one camera, the success rate decreases by 14.3\%-18.6\%, which has significant implications for real-world attack scenarios. When visible in two or three cameras, the attack's effectiveness increases (24.7\%-31.2\% and 31.2\%-39.9\%, respectively). Notably, the attack is more effective from the wrist cameras (16.2\%-21.3\%) than from the top camera (11.8\%-14.9\%), likely because wrist cameras provide more critical visual information for task execution. These results show that the adversarial object does not require precise control over its appearance across multiple viewpoints to be effective, validating its utility in dynamic real-world environments.

\subsection{Real-world Validation}
\label{sec:real_world}

To validate the attack in a real-world physical environment, we selected five representative tasks and deployed the adversarial patches optimized in simulation onto the Aloha dual-arm robot platform. The UAO was 3D printed into a 3cm radius sphere, with the optimized texture applied. Table \ref{tab:real_world} shows the sim-to-real transferability of the attack. Despite challenges such as printing inaccuracies, dynamic lighting, and camera calibration errors, the attack still maintained significant effects. The success rates for RDT and Pi0 models dropped by 81.4\% and 82.2\%, respectively, compared to simulation, demonstrating the robustness of our adversarial object to real-world noise. In particular, tasks of varying complexity exhibited similar transfer rates (79\%-84\%), confirming the robustness of our attack method.

\begin{table}[!t]
\centering
\caption{Real-world validation of the adversarial attack on physical Aloha robot platform. Transfer Rate indicates the percentage of attack effectiveness retained from simulation.}
\label{tab:real_world}
\resizebox{\columnwidth}{!}{%
\begin{tabular}{l|ccc|c|ccc|c}
\toprule
\multirow{2}{*}{\textbf{Task}} & \multicolumn{4}{c|}{\textbf{RDT}} & \multicolumn{4}{c}{\textbf{Pi0}} \\
\cmidrule{2-5} \cmidrule{6-9}
& \textbf{Clean SR} & \textbf{Adv. SR} & \textbf{$\Delta$SR↓} & \textbf{Transfer Rate} & \textbf{Clean SR} & \textbf{Adv. SR} & \textbf{$\Delta$SR↓} & \textbf{Transfer Rate} \\
\midrule
Grab Roller & 55\% & 23\% & 32\% & 82.1\% & 76\% & 36\% & 40\% & \textbf{83.3\%} \\
Adjust Bottle & 65\% & 28\% & 37\% & 82.2\% & 72\% & 35\% & 37\% & \textbf{84.1\%} \\
Pick Dual Bottles & 28\% & 10\% & 18\% & 81.8\% & 45\% & 20\% & 25\% & \textbf{80.6\%} \\
Place Burger Fries & 38\% & 15\% & 23\% & 79.3\% & 62\% & 28\% & 34\% & \textbf{81.0\%} \\
Press Stapler & 30\% & 12\% & 18\% & 81.8\% & 46\% & 23\% & 23\% & \textbf{82.1\%} \\
\midrule
\textbf{Average} & 43.2\% & 17.6\% & 25.6\% & 81.4\% & 60.2\% & 28.4\% & 31.8\% & \textbf{82.2\%} \\
\bottomrule
\end{tabular}
}
\end{table}

\section{Conclusion}
\label{conclusion}

In this work, we provide the first systematic analysis of the vulnerabilities in recent VLA models under adversarial attacks. By introducing a universal adversarial object attack method, we highlight that even state-of-the-art VLA models are susceptible to significant security risks. Our proposed multi-level attack framework demonstrates strong generalization across diverse tasks and models, achieving successful transfer from simulated environments to real-world scenarios.

In the future, we will focus on both offensive and defensive strategies. On the offensive side, further exploration of the various attack modalities can help uncover additional vulnerabilities in VLA systems. On the defensive side, developing robust defense mechanisms, including adversarial training and real-time detection systems, will be crucial to safeguarding the security and reliability of VLA technologies.

\section*{ACKNOWLEDGMENT}

\noindent This work was supported by the Key Research Project of Wuhan City 2024060788020073. The Learning Algorithms \& Soft Manipulation Laboratory of Wuhan University supported the robot in this paper.

\bibliographystyle{IEEEtran}
\balance 
\bibliography{root}

\end{document}